# From Stoplights to On-Ramps: A Comprehensive Set of Crash Rate Benchmarks for Freeway and Surface Street ADS Evaluation

John M. Scanlon[1, *], Timothy L McMurry[1], Yin-Hsiu Chen[1], Kristofer D. Kusano[1], Trent Victor[1]

[1] Waymo, LLC

* Corresponding author

## Abstract

This paper presents crash rate benchmarks for evaluating US-based Automated Driving Systems (ADS) for multiple urban areas, distinguishing between freeway and surface street crash rates, and breaking them down by crash severity and type. The purpose of this study was to extend prior benchmarks focused only on surface streets to additionally capture freeway crash risk for future ADS safety performance assessments. Using publicly available police-reported crash and vehicle miles traveled (VMT) data from Arizona, California, Georgia, and Texas, the methodology details the isolation of in-transport passenger vehicles, road type classification, and crash typology. Key findings revealed that freeway crash rates exhibit large geographic dependence variations with *any-injury-reported* crash rates being nearly 3.5 times higher in Atlanta (2.4 IPMM; the highest) when compared to Phoenix (0.7 IPMM; the lowest). The results show the critical need for location-specific benchmarks to avoid biased safety evaluations and provide insights into the vehicle miles traveled (VMT) required to achieve statistical significance for various safety impact levels. The distribution of crash types depended on the outcome severity level. Higher severity outcomes (e.g., fatal crashes) had a larger proportion of single-vehicle, vulnerable road users (VRU), and opposite-direction collisions compared to lower severity (police-reported) crashes. Given heterogeneity in crash types by severity, performance in low-severity scenarios may not be predictive of high-severity outcomes. These benchmarks are additionally used to quantify at the required mileage to show statistically significant deviations from human performance. Future work investigating the underlying factors influencing crash rates in each geographical area will further enhance future benchmarking efforts (by identifying potential confounders to account for when matching exposure between baseline and ADS data). This is the first paper to generate freeway-specific benchmarks for ADS evaluation and provides a foundational framework for future ADS benchmarking by evaluators and developers.

## Introduction

### Measuring ADS Effects on Traffic Safety

Injury and fatality risk continues to burden US roadways. Early estimates for 2024 indicate US traffic fatalities dropped (year-over-year) by 3.8% to 39,345 [1]. After a sharp rise in fatality rates

(the number of fatally injured persons per vehicle miles traveled, VMT) beginning in 2020, the fatality rate is beginning to drop back down to pre-COVID levels (before 2020). Looking at the last decade of available data (2013 - 2024), fatality rates have remained largely unchanged.

Stakeholders - the public, evaluators, regulators, and developers - are watching closely as fleets of SAE level-4 automated driving systems (referred to as ADS in this paper) are being deployed on public roadways [2]. An ADS is fully in control of the operation of the vehicle, and no human - present or remote - can take over the dynamic driving task. One primary use case being targeted by this technology is passenger vehicles operating as a rideshare service, which have the opportunity to replace a notable proportion of human-driven VMT with ADS-driven VMT at scale [3].

These deployments introduce the research question: do ADS-equipped vehicles have reduced injury and fatality risk on the roadways where they operate? This question is central to retrospective safety impact assessment, whereby the historical safety performance of some safety application is assessed through evaluating the in-field performance of the technology [4, 5]. These historical safety performance assessments have been commonplace in evaluating the on-road efficacy of vehicle safety interventions for many decades, and serve as key indicators for the true efficacy of the deployed technologies [6, 7, 8, 9, 10, 11, 12, 13]. ADS crash risk is most often historically measured by their crashed vehicle rate (sometimes referred to as a vehicle-level crash rate), which is the frequency that a vehicle crashes per VMT (often referred to as a "crash rate"). An identical metric to the crashed vehicle rate is the "driver-level crash rate", which is computed by counting the number of times a set of drivers crash per VMT. Baseline crash risk, sometimes referred to as a benchmark, is also central to this risk calculation, and is taken by counting the number of baseline vehicles that have crashed per VMT [5]. A basic form of the safety impact measurement can be seen in equation 1, below.

[1] ADS crash rate percent difference to benchmark (Safety impact) = (ADS Crash Rate / Baseline Crash Rate - 1) %

= percent difference in crash risk (negative value indicates a lower ADS crash rate)

Safety impact has only more recently been applied to evaluating ADS deployments. Early work examined testing operations (with a human operator in the vehicle or remote), while more recent studies have begun evaluating the safety impact of fully deployed ADS operations (without a human operator; sometimes referred to as rider-only) [13, 14, 15, 16, 17, 18]. This early flurry of studies prompted a group of traffic safety experts, including academics, developers, and insurance groups evaluating this technology, to create a working group to identify challenges and best practices for performing this work. The outcome of this collaboration was the RAVE Checklist [4] that details a number of steps to be taken when designing, implementing, documenting, and evaluating retrospective safety impact studies. The current benchmarks paper is intended to help support future studies aiming to conform to these best practices by (among others, but most notably):

- Providing a framework for using publicly-accessible data.
- Clearly documenting a methodological approach that can be replicated.

- Quantifying the effects of reducing analytical bias due to geographical, vehicle, and road type effects.
- Providing a wide set of crash type and severity benchmarks for enabling a diverse set of assessment scopes.

## Defining and Developing Crash Rate Benchmarks

Benchmark creation for the purposes of measuring ADS safety impact is a developing research area, and a limited set of crash rate benchmarks have been created over the past decade. Some of the biggest challenges identified in the RAVE Checklist are: (1) accounting for potential exposure differences that might bias the data, (2) differing outcome levels (e.g. crash severity levels) and reporting practices between benchmark and ADS, (3) analyzing the data through multiple analytical lenses, and (4) enabling reproducibility. Each of these challenges can be partially addressed by using publicly-accessible VMT and police-reported crash data to generate benchmarks.

Before describing the benchmarks currently available, it is important to first recognize what ADS data is currently available, or possible, for comparison. To do a comparison, researchers must reasonably match both exposure and crash outcome data to facilitate an apples-to-apples evaluation [4]. ADS-equipped vehicles are continuously collecting data from their vehicles about where and when they operate. This facilitates granular *exposure data* analysis of the VMT being accumulated, such as when and where they are operating. ADS providers are not required to report exposure data, but at least one ADS provider, to date, has routinely disclosed this information to facilitate external analysis [19]. Second, ADS-equipped vehicles are required in the US under the NHTSA Standing General Order (SGO) 2021-01 to report crashes that involve any amount of property damage or injury. The SGO data captures information about whether police are investigating or an injury was known to have occurred. One ADS operator has, to date, provided supplemental outcome information to supplement the SGO and enable more benchmark comparison points, including the documentation of (1) whether a suspected serious injury occurred or (2) a police report was filed.

Any benchmarks created should focus on aligning exposure and outcome data according to what can be measured from the ADS data (provided that the ADS operator released the necessary VMT to make a comparison). Table 1 provides a current accounting of research studies that have generated crash rate benchmarks for US driving. The table contains a number of features currently incorporated into available benchmarks, but what is not shown are the potential features that can be added in the future given their potential to (a) further account for confounders affecting crash risk and (b) examine the crash data through additional analytical lenses. Benchmark features needing further exploration are presented in the discussion limitations section "Unaccounted for Effects and Explainability of the Crash Rates Variation."

**Table 1.** A compilation of prior studies that have generated benchmarks for ADS safety performance assessment.

| Study | Crashed Vehicle Rates[1] | Driver Population | Geographic Area | Temporal Effects | Vehicle Type | Road Type | Outcome Levels | Crash Type Variants |
|---|---|---|---|---|---|---|---|---|
| *Insurance Claims Data* | | | | | | | | |
| Di Lillo et al., 2024 [16] | ✔ | garaged zip code | Residence Zip-Code | ✘ | pass. vehicles | All roads (Combined) | Property + Injury Claims | ✘ |
| Di Lillo et al., 2025 [17] | ✔ | garaged zip codes w/ latest vehicles | Residence Zip-Code | ✘ | pass. vehicles | All roads (Combined) | Property + Injury Claims | ✘ |
| *Naturalistic Driving Study Data* | | | | | | | | |
| Blanco et al., 2016 [21] | ✔ | US Population | Entire US | ✘ | pass. vehicles | All roads (Combined) | Police reportable | ✘ |
| Teoh and Kidd, 2017 [31] | ✔ | Local Population | MTV, CA | ✘ | pass. vehicles | surface streets | Police reported w/ subsets | ✔ |
| Goodall et al., 2021 [26] | ✔ | All Drivers | Entire US | ✘ | pass. vehicles | All roads (Combined) | Police reportable | ✔ |
| Goodall et al., 2021 [27] | ✔ | "Model" + All Drivers | Entire US | ✘ | pass. vehicles | All roads (Combined) | Police reportable | ✘ |
| Flannagan et al., 2023 [30] | ✔ | Rideshare Drivers | San Francisco | ✘ | pass. vehicles | All roads (Combined) | Any property damage | ✘ |
| *Fleet Reported* | | | | | | | | |
| Cummings, 2024 [14] | ✔ | Rideshare Drivers | California | ✘ | pass. vehicles | All roads (Combined) | Self-reported crashes | ✘ |
| Chen and Shladover, 2024 [15] | ✔ | Rideshare Drivers | California | ✘ | pass. vehicles | All roads (Combined) | Self-reported crashes | ✘ |
| *Police-Reported + Public Mileage Data* | | | | | | | | |
| Schoettle and Sivak, 2015 [25] | ✘ | All Drivers | Entire US | ✘ | All types[2] | All roads (Combined) | Police reported | ✘ |
| Banerjee et al., 2018 [20] | ✘ | All Drivers | Entire US | ✘ | All types[2] | All roads (Combined) | Police reported | ✘ |
| Blanco et al., 2016 [21] | ✘ | All Drivers | Entire US | ✘ | All types[2] | All roads (Combined) | Police reported | ✘ |
| Dixit et al., 2016 [23] | ✘ | All Drivers | Entire US | ✘ | All types[2] | All roads (Combined) | Police reported | ✘ |
| Favarò et al., 2017 [24] | ✘ | All Drivers | Entire US | ✘ | All types[2] | All roads (Combined) | Police reported | ✘ |
| Teoh and Kidd, 2017 [31] | ✔ | All Drivers | MTV, CA | ✘ | pass. vehicles | surface streets | Police reported w/ subsets | ✔ |
| Cummings, 2024 [14] | ✘ | All Drivers | Entire US | ✘ | All types[2] | Non Interstates | Police reported | ✘ |
| Scanlon et al., 2024 [32] | ✔ | All Drivers | Multiple counties, Entire US | ✘ | pass. vehicles | All, surface streets | Police reported w/ subsets | ✘ |
| Chen et al., 2024 [33] | ✔ | All Drivers | Dynamic matched | Time-of-Day | pass. vehicles | Surface Streets | Police reported w/ subsets | ✘ |
| Kusano et al., 2025 [19] | ✔ | All Drivers | Dynamic matched | ✘ | pass. vehicles | Surface Streets | Police reported w/ subsets | ✔ |
| Scanlon et al., 2025 (current) | ✔ | All Drivers | Multiple counties | ✘ | pass. vehicles | Surface Streets & Freeways | Police reported w/ subsets | ✔ |

[1] The current study focused exclusively on crashed vehicle rates, the number of crashed vehicles per VMT. Studies that did not present crashed vehicle rates are highlighted in red. Mixing crashed vehicle rates (the units when measuring ADS performance) with some non vehicle-level crash rate benchmarks creates a unit

mismatch, which prevents a valid mathematical comparison (see RAVE Checklist recommendation 1C in [4]).

[2] All vehicles include passenger vehicles, motorcycles, heavy vehicles, recreational vehicles, and other vehicles (e.g., crashes involving golf carts).

There has been a steady progression in complexity of benchmarks, as a result of increased precision. The simplest, most repeated approach leveraged readily available US police reported crash counts and VMT to broadly estimate the overall driving crash risk across the US without consideration of specific geographic area, vehicle type, or road type [20, 21, 23, 24, 25]. Cummings [14] performed a version of this national crash rate analysis but excluded interstate driving (leaving in other freeways and expressways that are also generally high speed and access-controlled) to focus on lower speed roadways that are more representative of actively deployed fleets. Under all these broad analyses of US national crash rates, police reported benchmarks were created for human drivers that historically are well known to have a large degree of underreporting, which biases the comparison to ADS data that has a more stringent reporting requirement [53].

National crash rate estimates have also been generated multiple times using SHRP-2 Naturalistic driving study (NDS) data [21, 26, 27, 28]. SHRP-2 was designed to be approximately representative of the entire US driving fleet [22]. Blanco et al. analyzed SHRP-2 by multiple severity outcome levels (including a police reportable level that negates effects of underreporting but introducing subjectivity of what is reportable) and road types (including interstate roads that had a crash rate 20% of the urban driving level) [21]. Goodall analyzed a subset of SHRP-2 crashes to generate front-to-rear struck crash rates across the entire set of geographic areas (again, intended to be representative of the entire US) [26]. In another study, Goodall looked at how police report data and NDS benchmarks could be adjusted to reflect "model driving" (defined to represent "sober, rested, attentive, and cautious" individuals that crash at two-thirds the rate of the entire population) [27].

More geographic area focused naturalistic driving studies have also taken place, but none have quantified freeway crash rates. Cummings [14] and Chen and Shladover [15] both developed crash rate benchmarks representative of rideshare drivers on all road types in San Francisco using California Public Utilities Commission (CPUC) Transportation Network Companies (TNC) reporting. The CPUC TNC does not have standardized reporting requirements making it likely that minor crashes are not fully reported nor is the crash outcome (e.g., property damage, police report, injury) reported. These limitations in the TNC CPUC data makes it impossible to align crash outcome levels with ADS data for a comparison [4, 29]. Flannagan et al. [30] developed San Francisco crash rate benchmarks using 5.6M VMT of data collected from an instrumented ridehailing fleet operating on surface streets. An instrumented fleet enabled Flannagan et al. [30] to capture many lower severity crashes. However, similar to the reporting issues with CPUC TNC data, not enough information about property damage and associated injuries (or other severity measures) were available to compare to ADS driving.

Third party insurance liability claims have more recently been introduced as a benchmark, which have more standardized reporting practices. Di Lillo et al. [16, 17] generated passenger vehicle crash rate benchmarks using third party liability claims data from residents for multiple individual geographic areas. The authors did not subset their crash rates by road type (instead using all roads combined).

Only a small set of studies have looked at geographic specific crash rates using police reported data. Blanco et al. [21] computed crash rates using police-reported data across all roadways for multiple specific geographic areas. Teoh and Kidd [31] were the first to compute roadway, vehicle, and geographic specific benchmarks using police reported data, where they targeted surface streets in Mountain View, California. Scanlon et al. [32] similarly targeted ADS deployment areas on surface areas using location specific police reported data, while also selecting surface streets and passenger vehicles. Chen et al. [33] developed dynamic spatial and temporal adjustments for the surface street benchmarks in Scanlon et al. [34] by accounting for when and where the ADS being benchmarked is operating within some deployed region. Kusano et al. [19], in an extension of [33, 34], further added in crash type distributions for more granular analysis of the performance.

## Emerging Operations on Freeways

The currently established benchmarks have all focused on either all roads or surface street driving with the exception of the Blanco et al [21] that quantified a SHRP-2 interstate crash rate (without further geographic limitations). Freeway driving conditions are considerably different from surface streets. This paper defined "freeways" using the Federal Highway Administration (FHWA) functional classifications for "interstates" and "principal arterials: other freeways and expressways" [34]. Freeways are designed to handle high speed movement of a large volume of vehicles along set corridors [35]. These roads tend to have higher speeds, be ramp access-controlled (no intersections), have lower exposure to pedestrians and cyclists, have barriers dividing opposite flowing traffic, among many other design characteristics. The remainder of roads are referred to as "surface streets", which include local, collector, and non-freeway arterial roads. These roads have lower speeds, higher exposure to vulnerable road users, and intersections (which introduces cross traffic crashes; a major contributor to US road injury burden [36, 37]). These differences have been found to create differences in crash risk. For example, using 2023 national fatal statistics (from the Fatality Analysis Reporting System, FARS) and associated vehicle miles traveled (VMT) estimates, the Federal Highway Administration's (FHWA) annually released fatality rate statistics as a part of their highway statistics series [38, 39]. In 2023, national fatality rates in the US were found to be 2.7-times higher on surface streets than freeways [39] (see Appendix A for more information).

As of this writing, the vast majority of ADS operations measured by VMT are on US surface streets. Given the early stages of ADS freeway operations (no passenger vehicle commercial operations), it is unsurprising that the existing benchmarks (see Table 1) have not directly quantified freeway crash risk since these existing studies are both generating benchmarks and performing a comparison to some existing operation. As multiple companies are preparing to deploy ADS in the freeway environment, freeway benchmarks developed in this paper are important to specify to enable future safety impact evaluations.

## Objective and Research Question

The purpose of this current study is to generate freeway crash rate benchmarks for the purposes of evaluating future ADS deployments. This study additionally provided updated surface street benchmarks (later crash year, updated road type classification methodology using new data) in order to contextualize the freeway benchmarks. This analysis represents an expansion of the work by Scanlon et al. [32] and Kusano et al. [19], whereby passenger vehicle, surface street crash rates were generated from police-reported data for multiple urban cities. We aimed to create benchmarks having the following attributes:

(1) Leveraging publically-accessible data that can be replicated by third-parties,
(2) Crashed vehicle rates,
(3) Geographic-specific (targeting multiple urban US counties),
(4) Representative of in-transport passenger vehicles,
(5) Separated by freeway and surface street (updated surface street definitions from prior work using new data),
(6) Separated by driving year,
(7) Broken down by multiple, common crash types, and
(8) Extensible to further incorporate dynamic adjustments (e.g., spatial and temporal effects; Chen et al. [33])

Multiple research questions were pursued.

(1) How do freeway driving crash rates compare to surface streets?
(2) How does geographic location influence those crash rates?
(3) How do crash rates vary by crash severity levels?
(4) What is the distribution of crash types?
(5) How many miles are needed to establish statistical significance?

# Methodology

## Data Source

Three sets of data were relied upon in the current study to source information on crash frequency and driven VMT (used to compute crash rates) for the various geographic locations. All crash and mileage data were taken from the year 2023, and represent the totality of that year. First, police reported crash datasets were obtained from individual states to establish crash frequency. The states provided the data in table form at the crash, vehicle, and person levels. All of the crash data is publicly accessible. It is notable that some states require a request for the data (with contact information being provided on the respective crash data websites). Second, vehicle miles traveled (VMT) were obtained from each state's reporting. For some states, the VMT estimates were available by road type and were relied upon. Third, California did not have breakdowns of VMT by road type (i.e., they only reported overall VMT). The Federal Highway Administration (FHWA) maintained Highway Performance Monitoring System (HPMS) data was used to directly determine the VMT on Freeways by county in

California. Arizona, Georgia, and Texas provided county-level VMT by functional classification, and did not require this additional step using HPMS.

**Table 2:** A breakdown of the primary mileage data sources relied upon in the current study to quantify crash rates.

| State | Crash Data | Mileage Source | | |
|---|---|---|---|---|
| | | All Mileage | Freeway | Surface |
| Arizona | Arizona Department of Transportation (ADOT) [40, 41] | AZ Certified Public Miles [42] | | |
| California | California Highway Patrol California Crash Reporting System (CCRS)* [43] | CA Public Road Data [44] | HPMS [49] | All - HPMS Freeway |
| Georgia | GDOT GEARS crash data [45] | GDOT Road Mileage Reports [46] | | |
| Texas | TXDOT Crash Records Information System (CRIS) [47] | TX Transit Statistics [48] | | |

*Previously named the Statewide Integrated Traffic Records System (SWITRS).

## Data Features

### Crashed Vehicle Rates

We created crashed vehicle rates (vehicle-level crash rates), defined as the number of crashed vehicles per VMT. These rates are created by simply dividing the *number of crashed vehicles* by the driven VMT. For compliance with RAVE checklist 1C [4], this approach ensures a direct comparison to ADS performance, where ADS crashes are commonly evaluated by the number of times ADS vehicles crash relative to the amount of driven VMT. Unit mismatch errors have been commonly observed in ADS benchmarking studies, where non vehicle-level crash rate benchmarks have been compared to ADS crashed vehicle rates (see Table 1) [4].

### Geographic-Specific

We generated crash rates for multiple counties as well as aggregated rates for wider service areas (e.g. metro Atlanta). The geographic areas and their associated definitions can be found in Table 3.

**Table 3:** Geographic areas investigated in the current study.

| Geographic Area | State | County |
|---|---|---|
| Atlanta | Georgia | Fulton |
| Atlanta | Georgia | DeKalb |
| Atlanta | Georgia | Clayton |
| Austin | Texas | Travis |
| Los Angeles | California | Los Angeles |
| Phoenix | Arizona | Maricopa |
| San Francisco to San Jose | California | San Francisco |
| San Francisco to San Jose | California | San Mateo |
| San Francisco to San Jose | California | Santa Clara |

### In-Transport Passenger Vehicle Type

Crash rates were specific to in-transport (e.g., not parked), passenger vehicles. The passenger vehicle classification aimed to reflect a gross vehicle weight rating (GVWR) less than 10,000 pounds (49 CFR § 565.15) [50]. Scanlon et al. [32] previously found that including all vehicle types, e.g., heavy trucks and motorcycles, in the crash rate computation led to a slight underestimation of passenger vehicle only crash risk (approximately 5% using national crash data).

Data source specific methodology was used to select passenger vehicle mileage and crash data. Appendix Table A1 lists the variables relied upon to subselect passenger vehicles in the crash data. For the crash data, passenger vehicles in-transport were first isolated using the state-specific vehicle type variables available. For a small proportion of vehicles, the specific vehicle type could not be determined (e.g., when a party left the scene of a crash before police could determine the vehicle type). These unknown vehicle types were imputed using the distribution of known vehicle types at the geographic level (reflecting likelihood of that vehicle being a passenger vehicle) [32].

For the mileage data, VMT was not provided at the vehicle-type level. To determine the proportion of the total mileage attributable to passenger vehicles, the FHWA Highway Statistics Series tables (specifically VM4) were used, which lists out the proportion of VMT attributed to various vehicle types by state, road type, and urban / rural designation [38].

### Road Type

As noted previously, this study quantified crash risk by surface street (FHWA functional classifications: local, collectors, and non-freeway arterials) and freeway (interstates and "principal arterials: other freeways and expressways") road types. The current methodology updated prior work by Scanlon et al. [32]. Previously, surface streets were identified using a combination of state data specific variables, including road name and speed limit, with the goal of having high recall of surface streets. The new approach added in the usage of HPMS data, which contains freeway classifications and mileage, by year, for all of the states being analyzed.

The encoded crash latitude, longitude, and road name were used to match the appropriate FHWA functional classification from HPMS classification to the crash data. The road classification in the crash data occurred in two basic steps. The first step was to determine whether the associated road name corresponded to a known freeway road type. Naming conventions are consistent across all state databases, and can be used to identify various freeway road names that might be interstates, state routes, or US routes. For example, "I-280 N/B" in San Francisco county is readily identifiable as interstate 280. There are also more localized names to identify, such as Texas Loop 1 being locally called (and coded as) "MOPAC". Compared to surface streets, there are relatively few freeway variants in each of the listed geographic areas, which could be directly identified using regular expression. To ensure completeness of the naming classifier, the HPMS "route_id" variable was used to search for all city-specific road name variants. Second, some of the identified freeway names can change functional classification along the road. For these road names associated with both surface streets and freeways, the next step in the process was to examine whether the identified crash was along the known stretch of freeway. This was accomplished by comparing the crash location provided in degrees latitude and longitude to the nearest HPMS freeway (looking for a proximity of 400 m to the known freeway). As an example, I-280 in San Francisco is always a freeway, and would not require this additional step. On the other hand, US Route 101 is both a freeway in some parts then transitioning to a surface street, so the second proximity to freeway evaluation was needed to determine where on the road the crash happened.

For California SWITRS data, latitude and longitude were not always encoded. As done in Chen et al. [33], missing coordinates were determined using Google's geocoding API by inputting the known crash city ("CNTY_CITY_LOC") and involved roadways ("PRIMARY_RD" and "SECONDARY_RD").

### Crash Type

Various crash typology has been leveraged for grouping and describing pre-crash perspectives [51, 52]. This study relied on the groupings described in Kusano et al. [19], which are shown in Figure 1. The variables used to do this classification were specific to the state crash database being utilized and are listed out in the Appendix (Table A2).

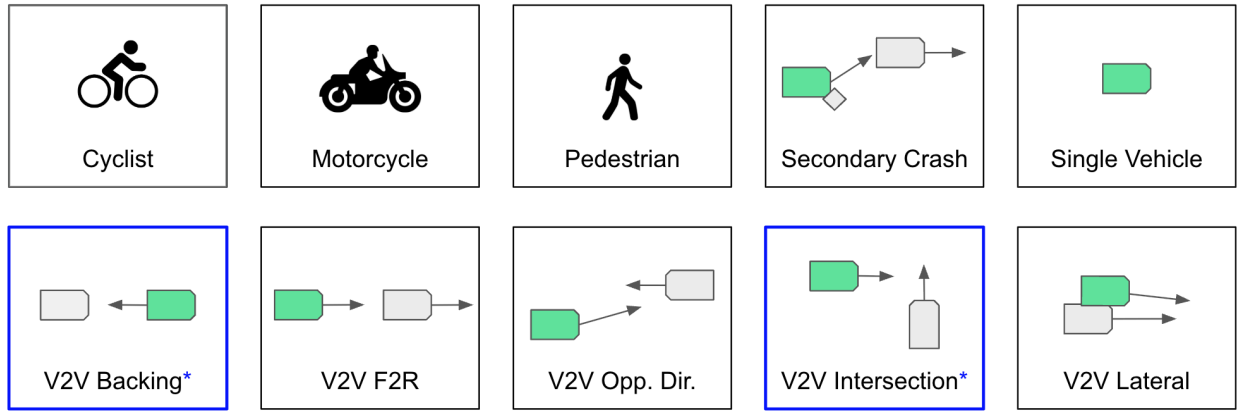

**Figure 1.** Crash types relied upon in the current study for breaking down crash rates.

This aggregation scheme relied upon in this study collapses the conflict typology presented in Kusano et al. [51]. The conflict typology has dimensions of collision partners, pre-crash maneuvering and road context, initiator and responder role, and causal mechanisms, and is designed to be collapsible and actionable for individual use cases. The crash types were intended to balance a classification that is informational (reflective of relevant causal and severity factors) and one that can be statistically evaluated under a reasonable deployment time horizon (e.g., hundreds of millions of miles). Notably, three vulnerable user groupings were presented due to their overrepresentation in urban environments and high severity potential. The remainder of groups broadly bucketed vehicle-to-vehicle and single vehicle interactions, which mostly align with geometric aggregations notably within NHTSA's pre-crash Scenario Typology [52]. Because freeway driving typically does not include cross traffic, the intersection grouping was only applied to surface streets. The "Secondary Crash" bucket captures collisions where the ego vehicle is not involved in the first contact event (but is involved in a second or later contact). An "Unknown / Other" group was relied upon for crashes without sufficient information for bucketing or for crashes that did not fit into one of these buckets.

### Outcome Levels

Multiple outcomes (Table 4) were considered that can be directly determined from police reported data. The variables used to select each outcome for each state database are listed in the Appendix. The outcome levels were previously described by Scanlon et al. [32].

**Table 4.** Description of Outcome Levels from Police-Reported Crash Data.

| Outcome Level | Description |
|---|---|
| *Police-Reported* | The event was reported by police. |
| *Any-Injury-reported* | An injury at any level was reported to have occurred. An underreporting adjustment was |

|  | applied using NHTSA estimates. |
|---|---|
| *Any Airbag Deployment* | Any vehicle throughout the entire collision sequence experienced an airbag deployment. |
| *Suspected Serious Injury+* | Any person throughout the entire collision sequence experienced a suspected serious injury (A-level) or fatality (K-level). |
| *Fatal* | Any person throughout the entire collision sequence experienced a fatal injury. |

*Police-reported* crashes are, by definition, all events included in the state crash databases. As noted in Scanlon et al. [32], there are many known challenges associated with underreporting and relying on the *police-reported outcome* level for safety impact analysis. These challenges can generally be broken down into: (a) what reporting thresholds are present for dictating what is police reportable? and (b) how likely is a reportable event to be reported? Both of these challenges limit the reliability of a *police reported* benchmark crash rate. The underreporting challenge is particularly noteworthy when considering prior work by Blincoe et al. that found 60% of property-damage-only and 32% of non-fatal injury crashes were not reported to police across the US. Another notable challenge is, in California, property damage only crashes (no injury) reported by police are not required to be entered into the state crash database. For completeness, the *police reported* crash rate is provided in this study, but, due to these aforementioned reasons, it is not recommended to be used for safety impact analysis. Instead, the authors recommend relying on the latter outcomes discussed in the subsequent paragraphs, which have more straightforward reporting thresholds.

*Any-Injury-reported, Suspected Serious Injury+,* and *fatal* crashes were all identified using the encoded "KABCO" injury classification at the crash-level [54]. If any individual involved in the crash sequence was injured (e.g., occupant, VRU, secondary collision partner), the crash-level KABCO classification will capture that worst injury. Each state has a unique KABCO classification scheme definition, so caution should be used when comparing non-fatal injury outcomes between states. *Any Airbag Deployment* crash rates look for the occurrence of an airbag in *any* of the involved vehicles. For example, if a crash sequence involved four vehicles and only one of those vehicles experienced an airbag deployment, every vehicle involved would receive the same *"any airbag deployment"* classification.

As has been done previously for *any-injury-reported* crash rates [21, 25, 32], the non-fatal proportion of injury crashes were adjusted using NHTSA estimates for underreporting [53]. No underreporting adjustment was performed for any of the other crash rates, including police-reported outcomes that are a superset of *any-injury-reported* crashes.

## Power Analysis Methodology

Power analysis was relied upon to estimate the amount of VMT required to determine statistical significance given assumed ADS crash rates relative to the human benchmark. Power analysis is a common technique employed in research design prior to data collection to determine the amount of data needed to test some hypothesis [55]. This specific statistical approach was previously used in Scanlon et al. for evaluating surface streets [32]. The VMT required to draw statistically significant conclusions in a power analysis is a function of the benchmark rate and effect size (i.e., relative ADS performance). Several ADS performance levels were evaluated, including 25%, 50%, 75%, and 90% reductions in crash rates and 25% and 50% increases in crash rates. Sample size formulas were used that model asymptotic normal approximations to a Poisson distribution. The method targeted 80% statistical power using two-sided tests with $\alpha$ = 0.05.

In detail, we calculated required mileage using the formula below, where $\lambda_{ADS}$ and $\lambda_{Human}$ are ADS and human crash rates, $\Phi^{-1}(.)$ is the standard normal quantile function, 1-$\beta$ is the desired statistical power (e.g. 0.8), and $\alpha$ is the probability of type I error (e.g. 0.05).

$$\text{Required Mileage} = \frac{\left(\sqrt{\lambda_{ADS}}\Phi^{-1}(1-\beta) + \sqrt{\lambda_{Human}}\Phi^{-1}(\alpha/2)\right)^2}{(\lambda_{ADS} - \lambda_{Human})^2}.$$

# Results

The complete set of benchmark rates are available for download as supplemental material to this paper. The following subsections pull from those benchmarks to examine each of the research questions.

## Freeway Crash Rates

The freeway crash rates are listed by geographic region and severity level in Table 5. There were clear geographic and severity level variations, where no two geographic regions had similar crash rates across all severity level variations. As noted in the methodology, there are geographic-specific underreporting, reportability, and outcome definitions to consider when making comparisons across geographic areas. To that end, consider any *airbag deployed* and *fatal* crash rates, where the former is over an order of magnitude more frequently occurring and the definitions are equivalent across all geographic regions. The *any airbag deployed* freeway crash rates varied from 0.355 (in Phoenix) to 1.033 (in Atlanta) incidents per million miles (2.9-times difference). Similarly, for *any fatal* freeway crashes, the rates varied from 5 (in Phoenix) to 14 (in Atlanta) incidents per billion miles (2.8-times difference).

**Table 5.** Freeway crash rates generated in the current study by geographic area and severity level using 2023 data.

| | Geo | Atlanta | Austin | Los Angeles | Phoenix | San Francisco to San Jose |
|---|---|---|---|---|---|---|
| | Counties | Fulton, DeKalb, Clayton | Travis | Los Angeles | Maricopa | San Francisco, San Mateo, Santa Clara |
| | Mileage (Mmi) | 10,180 | 4,279 | 30,700 | 31,285 | 9,965 |
| Crashed Passenger Vehicles (IPMM) | *Police-Reported* | 57,103 (5.609) | 6,722 (1.571) | 72,034 (2.346) | 48,501 (1.550) | 20,648 (2.072) |
| | Any-Injury-Reported | 23,398 (2.298) | 4,176 (0.976) | 36,445 (1.187) | 23,342 (0.746) | 9,562 (0.960) |
| | Any Airbag Deployment | 10,519 (1.033) | 3,157 (0.738) | 20,786 (0.677) | 11,122 (0.355) | 6,695 (0.672) |
| | Suspected Serious Injury+ | 812 (0.080) | 199 (0.046) | 1,093 (0.036) | 491 (0.016) | 337 (0.034) |
| | Any Fatality | 140 (0.014) | 47 (0.011) | 179 (0.006) | 169 (0.005) | 71 (0.007) |

## Comparison to Surface Streets

Figure 2 shows a crash rate comparison between freeways and surface streets for all severity outcomes. Surface street crash rates were considerably higher than freeways across all geographic regions and severity outcomes. As a demonstration, for *any-injury reported* crashes, surface street rates were 1.7- to 2.8-times the freeway crash rate. On the other end of the spectrum, surface street *fatal* crash rates were 1.2- to 5.3-times higher than freeway *fatal* crash rates.

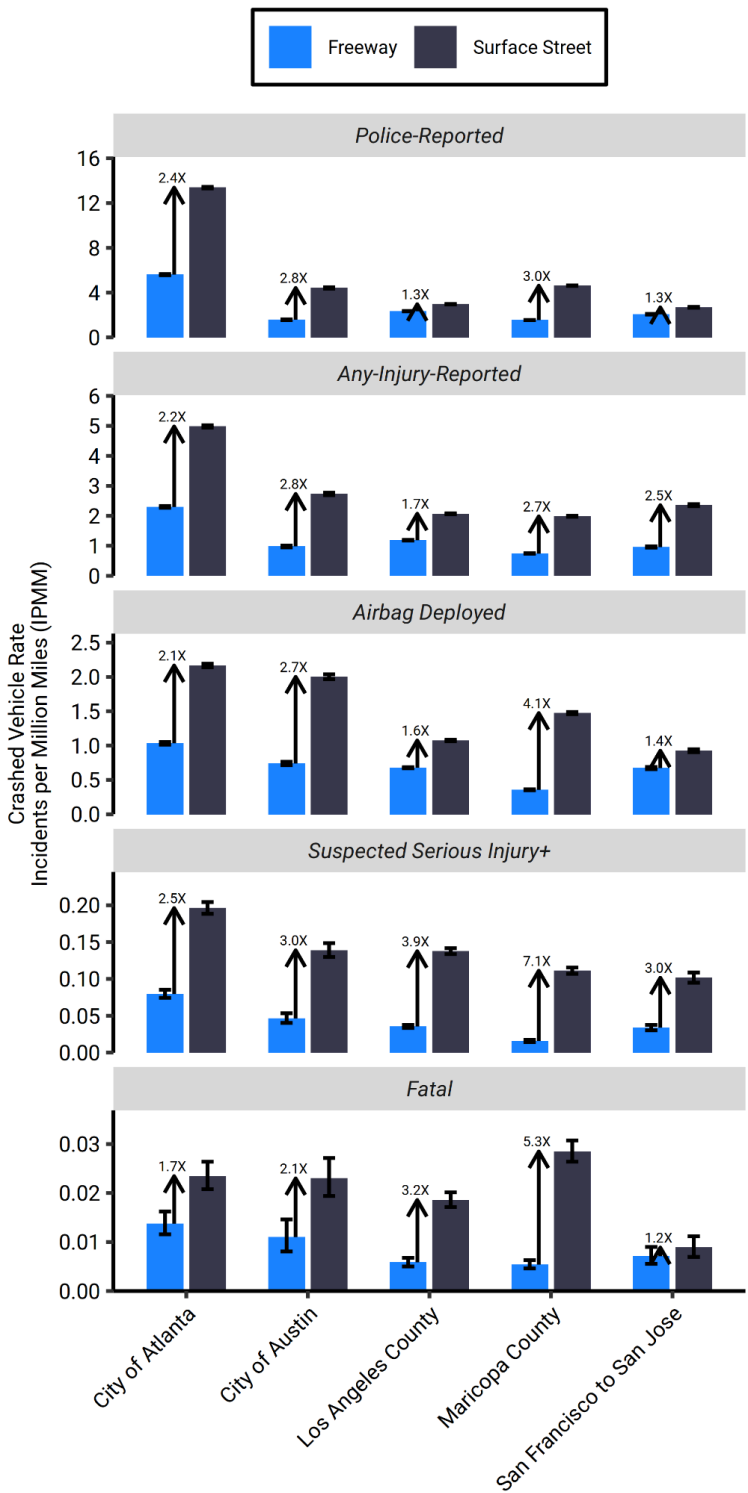

**Figure 2.** Crashed vehicle rate comparison between freeways and surface streets. The results are shown by geographic area and outcome level.

## Breakdown of Freeway Crash Types

Figure 3 shows the freeway crash types by geographic area and outcome level. Generally, each geographic region had similar distributions of crash types for given severity levels. There was, however, a notable trend in crash type distribution by severity level. Across all geographic areas analyzed, a few types of collisions made up the vast majority of *police-reported* crashes. Front-to-rear (45-53%) and vehicle-to-vehicle (V2V) lateral crashes (20-35%) together accounted for at least two-thirds of all police-reported incidents. When combined with single-vehicle (6-12%) and secondary crashes (8-18%), these four types comprised about 90% or more of the total. In contrast, the profile of crash types shifted substantially for *fatal* crashes. There was a notable increase in the proportion of crashes involving single vehicles (21-26%), vulnerable road users like pedestrians (9-21%) and motorcyclists (2-6%), and opposite-direction collisions (5-11%). Conversely, the relative share of V2V front-to-rear (9-23%) and V2V lateral (8-21%) crashes was lower for fatal crashes compared to police-reported crashes.

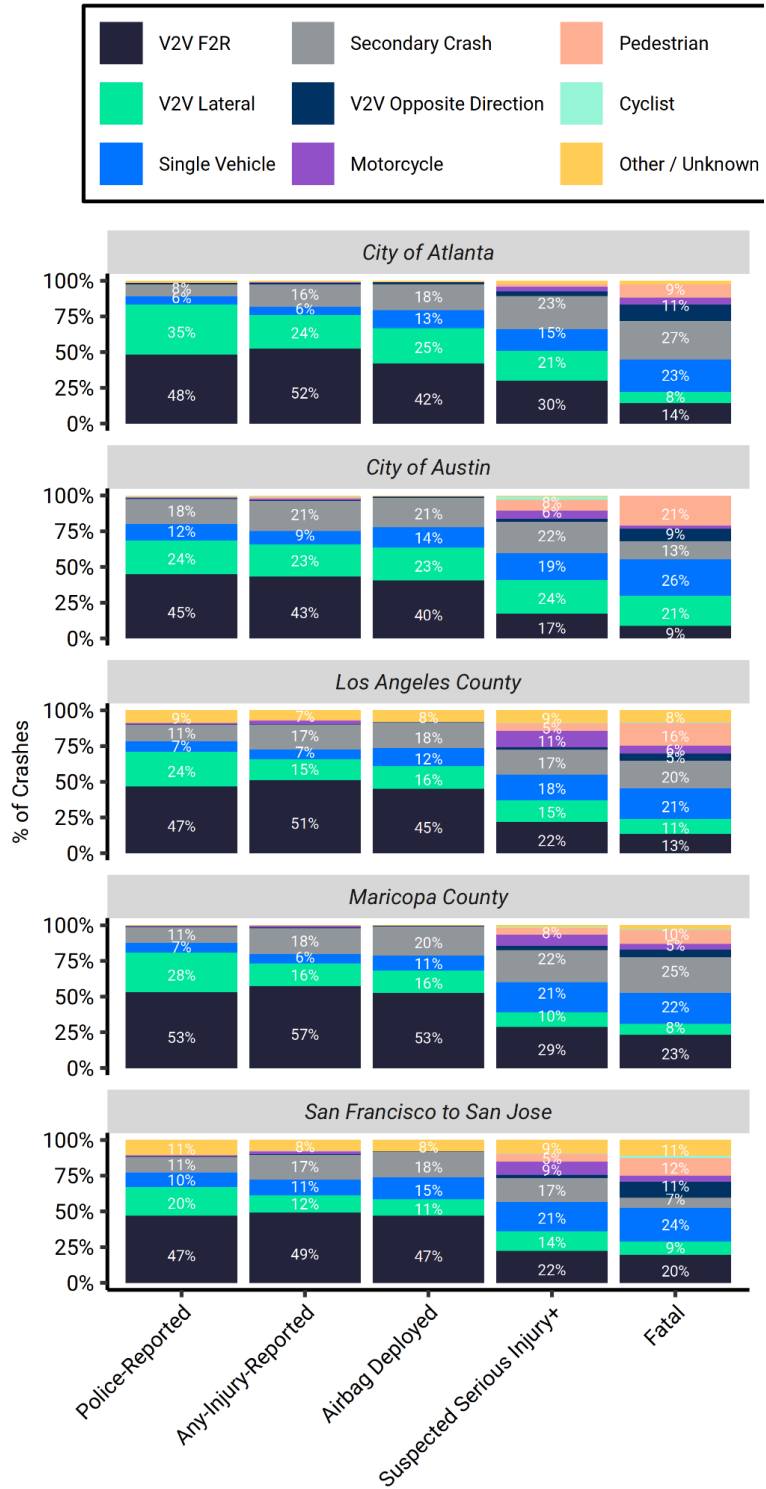

**Figure 3.** Crash type distributions are shown for each severity level and geographic area analyzed.

## Required Mileage for Statistical Testing

Figure 4 shows the vehicle miles traveled (VMT) required to determine the statistical significance of an ADS's performance across various road types, geographic areas, and severity levels. The complete set of required mileage is available for download in the supplemental materials. The required mileage is inversely proportional to the benchmark crash rate; a lower crash rate requires more data (mileage) to achieve statistical confidence. Therefore, because freeways have lower crash rates than their surface street counterparts, they require substantially more mileage to validate safety performance.

Furthermore, the required mileage decreases as the performance difference between the ADS and the human-driven benchmark grows. A larger safety improvement is easier to detect statistically. For example, to measure the safety impact on freeways:

With a **modest 25% safety improvement** (an ADS crash rate at 75% of benchmark), it would require:

- **21-75 million VMT** for police-reported crashes.
- **8.4-21.4 billion VMT** for fatal crashes.

With a **substantial 75% safety improvement** (an ADS crash rate at 25% of benchmark), the required mileage drops to:

- **1.8-6.5 million VMT** for police-reported crashes.
- **0.7-1.9 billion VMT** for fatal crashes.

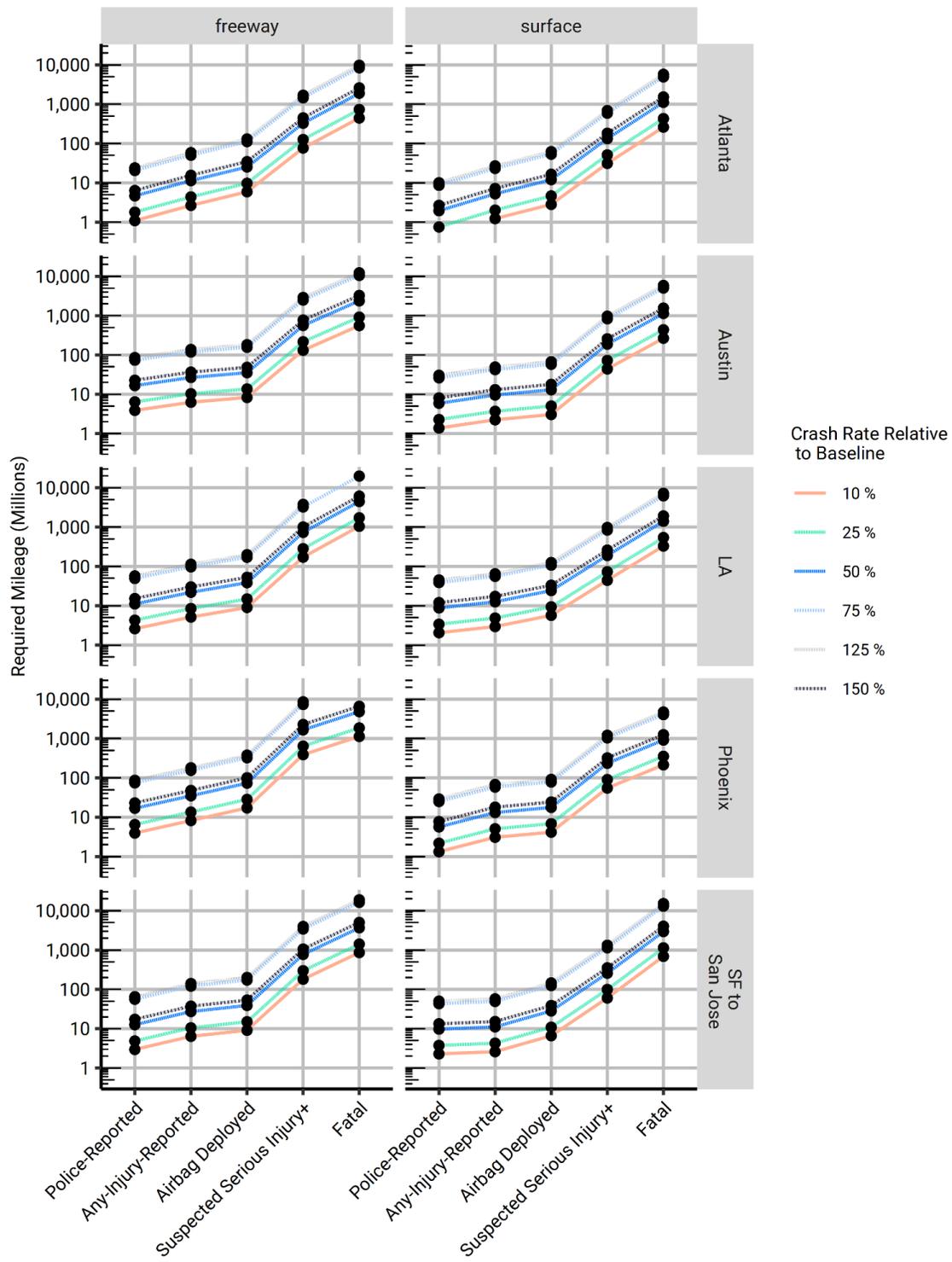

**Figure 4.** The required VMT to determine statistical significance is shown for each geographic area, road type, severity level, and performance level. The point represents the minimum amount of VMT required at that performance level.

# Discussion

## Implications of the Work

This work addresses a critical need for robust safety benchmarks against which ADS can be evaluated. We developed these benchmarks for freeway road types in key geographic areas where ADSs are already deployed (on surface streets), providing a direct, relevant baseline for comparison for future freeway deployments. A significant contribution of this research was that these benchmarks are derived entirely from publicly available data, ensuring the methodology is transparent and replicable. Furthermore, by disaggregating the data by crash severity and type, our work enables a more granular and meaningful assessment of ADS safety, moving beyond simplistic overall crash rates to understand specific performance benefits or shortcomings.

## Comparison to Surface Streets

A critical limitation in prior ADS safety benchmarking, as shown in Table 1, is the failure to differentiate between roadway types, often producing only a single "average" crash rate. This study addressed that gap directly by establishing distinct benchmarks for freeways and surface streets across multiple urban areas. Our results revealed a crucial finding: freeways have substantially lower crash rates than surface streets in all locations and at all severity levels.

These findings have major implications for future ADS benchmarking. Using a generic, all-road benchmark for freeway ADS driving creates a substantial exposure misalignment [4]. Benchmarks that are not specific freeways, but used to evaluate freeway driving, will *overestimate* - substantially - the safety impact of the ADS deployment. Conversely (and also demonstrated in Scanlon et al. [32]), if an all roadway benchmark was used to evaluate safety impact on surface streets, the safety impact of the ADS deployment would be *underestimated*.

Therefore, to avoid analytical bias, future ADS evaluations should use benchmarks specific to the roadway on which the system is operating. The methodology presented is publicly accessible and repeatable across multiple geographic areas.

## Geographic Effects on Freeway Crash Rates

Our analysis revealed that geographic location was a powerful indicator of freeway crash rates across all severity levels. This finding highlights a critical pitfall: using generalized, non-local crash rates can introduce non-negligible bias and lead to flawed safety conclusions. Consider the stark difference in fatal crash rates between Phoenix (4 incidents per billion miles) and Atlanta (15 IPBM). If an ADS operating in Phoenix exhibited a fatal crash rate of 15 IPBM, comparing it to Atlanta's benchmark would misleadingly suggest it performs on par with human drivers. However, when measured against the correct local benchmark, the analysis would reveal the ADS's fatal crash rate is over three times higher than that of average human drivers in Phoenix—a major safety concern that the wrong benchmark would completely obscure.

Two key factors influencing geographic variations are: (1) differences in crash reporting practices and (2) differing official definitions for each severity level. First, the likelihood of a crash being police-reported and included in the database varies by location and severity. National studies show substantial underreporting of police reported crashes, with an estimated 60% of property-damage-only and 32% of non-fatal injury crashes going unreported [53]. Local analyses can reveal even higher discrepancies. For example, a San Francisco Department of Public Health study comparing police and hospital records found that even for severe injuries (defined using an injury severity score (ISS) greater 15 or greater), a large percentage of injured vehicle occupants (39%), pedestrians (28%), and cyclists (33%) were absent from police reports [56]. Second, states may use different definitions for crash severity and have different levels of reportability. The minimum property damage value that requires a crash to be reported, for example, differs by state [53]. For this study, *any-injury-reported*, *any airbag deployment, suspected serious injury+, and fatal* crashes were found to have consistent definitions across the geographic areas in our study.

## Spatial and Temporal Effects within Geographic Area

This study provided crash rates at the county-level for a specific year of driving, which demonstrated a much needed level of granularity for ADS benchmarking. As freeway ADS mileage is accumulated, there is an opportunity to examine how different levels of spatial and temporal granularity affect these crash rates, such as crash rates for specific roadway segments within a geographic area or different temporal groupings (e.g., time-of-day, day of week, or season). Recent research by Chen et al. [33] demonstrated the importance of granular spatial and temporal factors on *surface street* ADS crash rate evaluation. Both the observation of underlying differences in ADS driving exposure and the dependency of crash risk on that exposure, motivated Chen et al. [33] implementation of a dynamic crash rate adjustment (to match driving exposure between baseline and ADS). It is also important to note the limitations of the available benchmarking data (see discussion section below), which often limit the type of adjustment that is even possible.

## Unaccounted for Effects and Explainability of the Crash Rates Variation

While this study accounts for reporting and definitional differences, one important next step would be to explore the underlying causal factors that drive geographic variations in crash risk. A vast body of traffic safety research indicates that factors such as traffic density, roadway features, urban / rural split, speed limits and prevailing speeds, weather patterns, driver and occupant demographics, seat belt use, and vehicle factors all play a role in crash frequency and severity [57, 58, 59, 60, 61, 62, 63, 64, 65]. However, the specific extent to which each factor contributes to risk in the studied geographic areas remains untested, and is a complex area of study given the amount of potential contributing factors. A deeper understanding of these influences would enable the creation of more powerful, dynamic benchmarks, whereby individual features could be better accounted for as confounders in an ADS evaluation. For instance, incorporating real-time variables like traffic flow or adverse weather could allow for adaptive risk models, dramatically enhancing the precision of ADS safety evaluations

## Severity-dependent Freeway Crash Types

The analysis of freeway crash types revealed a notable trend: the distribution of crash types varied according to the severity of the crash. At lower severity levels (e.g., all police-reported crashes), front-to-rear collisions and vehicle-to-vehicle lateral crashes constituted the majority. However, as crash severity increased to fatal outcomes, there was a discernible increase in single-vehicle crashes, vulnerable road user (VRU) involved crashes (specifically motorcyclists and pedestrians), and V2V opposite direction crashes (see figure 4).

This shift is expected, and reflects a tradeoff between frequency and severity. Freeway single vehicle, pedestrian, motorcycle, and V2V opposite direction, are all relatively less frequent police reported crash types, which likely indicates generally lower conflict exposure, but these events carry predictably high injury potential. Single vehicle crashes, such as road departures often involve high closing speeds with fixed and, often, narrow objects, considerable injury risk exists [66; 67; 68]. Pedestrian serious injury (AIS3+) probability is over 75% realm when the contacting vehicle is traveling over 60 mph [69], while motorcycle serious injury risk is nearly 50% at 60 mph [70]. Due to the nature of opposite direction crashes involving vehicles traveling in opposite directions, closing speeds have the potential to be highest, which accordingly translate to high collision forces and delta-v. The rarity and quickly diminishing "escape paths" make freeway oncoming vehicles inherently a more challenging scenario to anticipate and avoid or mitigate [71, 72, 73].

The observed heterogeneity in crash type distribution by severity will have implications on future ADS benchmarking evaluation. Put simply, the lower severity crashes (e.g., all *police-reported*) have a different distribution of crash types than higher severity crashes (e.g., *fatal).* These results suggest that ADS evaluators and developers should not assume that low severity crash performance will translate to high severity performance. For example, pedestrian crashes are relatively rare on freeways making up 1% or less of all *police-reported* crashes. An absence of any ADS freeway pedestrian crashes over only several 10s of millions of miles would not be a measurable statistical indicator of safety, because the benchmark data also has a relatively low crash rate. Conversely, when looking at future ADS *fatal* crash performance, pedestrian crashes make up 9-21% of the benchmark crashes, which indicates their relative importance at the *fatal* crash level for measuring safety impact.

The crash type heterogeneity further solidifies the RAVE checklist recommendation to "Exercise caution in relating findings between different severity levels" [4]. Specifically, safety impact performance in low severity events should not be assumed to be predictive of high severity events. The shift in crash type distribution reflects differences in underlying crash causation and avoidability [51]. In practice, ADS performance will first be evaluated against low severity outcomes. Even if performance in these lower severity outcomes is strong, the higher severity crashes will have a different distribution of crash geometries and collision partners, and ADS performance may be dependent on crash geometry and collision partners. This current analysis indicates that high severity freeway crash outcomes, at least in the human driven population, generally occur in rarer circumstances, which limits the transferability of any low severity performance to higher severity performance. Even with low exposure to ADS initiated collision scenarios, exposure to other initiators in high severity freeway conflicts, such as oncoming vehicles or pedestrians on freeways, is inevitable and necessitates careful, continuous evaluation as mileage is accumulated (and exposure to rarer circumstances grows) [74].

## Limitations of the Data

This study's reliance on publicly available crash and mileage data has clear advantages to facilitate replication and promote study transparency. Still, there are inherent limitations with these data sources. Crash data has a limited number of collected variables that are not easily comparable between datasets or to the available ADS data. More standardized crash reporting practices would help overcome these challenges. Additionally, the mileage data relied upon largely represents estimates by the states using samples of data over limited periods. The variance and bias due to these sampling methods have not been studied in this current study. Future research should (a) look to better examine these biases and (b) look for complementary data sources to help validate the existing data and enable more complex analyses.

# Conclusions

This study generated freeway crash rate benchmarks for evaluating ADS deployments on freeways within various urban US geographic areas. The data relied upon is all publicly available, and the methodology is extensible to new areas. It is strongly recommended that any implementation of these benchmarks explore how more granular temporal and spatial dynamic corrections can be used to better ensure apples-to-apples comparisons. The findings underscore the importance of location-specific benchmarks to avoid biased evaluations. The results emphasize that performance observed in lower-severity outcomes may not directly translate to higher-severity scenarios due to differing crash type distributions. The research also provides a foundational roadmap for future benchmarking studies, stressing the value of pre-specifying evaluation criteria for transparency and statistical rigor, and suggests that early, low severity statistical assessments are feasible even with limited freeway driving mileage.

# Acknowledgements

The authors would like to acknowledge the tremendous effort and support from each of the primary data source contributors. Your work to collect, organize, and disseminate data for safety research is a key enabler for transparent, public safety impact analysis that ultimately makes our roads safer. Thank you:

- Arizona Department of Transportation (ADOT)
- California Highway Patrol (CHP)
- California Department of Transportation (Caltrans)
- Georgia Department of Transportation (GDOT)
- Texas Department of Transportation (TXDOT)
- US DOT Federal Highway Administration (FHWA)

# References

1. National Highway Traffic Safety Administration. "Early estimate of motor vehicle traffic fatalities in 2024." Publication DOT HS 813 710 (2025).
2. SAE International Surface Vehicle Recommended Practice, "(R) Taxonomy and Definitions for Terms Related to Driving Automation Systems for On-Road Motor Vehicles," SAE Standard J3016, Revised April 2021.
3. Zeng, Teng, Hongcai Zhang, Scott J. Moura, and Zuo-Jun M. Shen. "Economic and environmental benefits of automated electric vehicle ride-hailing services in New York City." Scientific Reports 14, no. 1 (2024): 4180.
4. Scanlon, John M., Eric R. Teoh, David G. Kidd, Kristofer D. Kusano, Jonas Bärgman, Geoffrey Chi-Johnston, Luigi Di Lillo et al. "RAVE checklist: Recommendations for overcoming challenges in retrospective safety studies of automated driving systems." Traffic Injury Prevention (2024): 1-14, https://doi.org/10.1080/15389588.2024.2435620.
5. International Organization for Standardization (ISO). 2021. Road vehicles- Prospective safety performance assessment of pre-crash technology by virtual simulation - Part 1: State-of-the-art and general method overview. ISO/TR 21934-1.
6. Lie, Anders, Claes Tingvall, Maria Krafft, and Anders Kullgren. 2006. "The Effectiveness of Electronic Stability Control (ESC) in Reducing Real Life Crashes and Injuries." Traffic Injury Prevention 7 (1): 38–43. doi:10.1080/15389580500346838.
7. Glassbrenner D, Starnes M. 2009. Lives saved calculations for seat belts and frontal air bags (Report No. DOT HS 811 206). Washington,DC: National Highway Traffic Safety Administration.
8. Strandroth J, Rizzi M, Olai M, Lie A, Tingvall C. 2012. The effects of studded tires on fatal crashes with passenger cars and the benefits of electronic stability control (esc) in Swedish winter driving. AccidAnal Prev. 45:50–60. doi:10.1016/j.aap.2011.11.005.
9. Fildes B, Keall M, Bos N, Lie A, Page Y, Pastor C, Pennisi L, Rizzi M,Thomas P, Tingvall C., 2015. Effectiveness of low speed autonomous emergency braking in real-world rear-end crashes. Accid Anal Prev.81:24–29. doi:10.1016/j.aap.2015.03.029.
10. Isaksson-Hellman I, Lindman M. 2015. Evaluation of rear-end colli-sion avoidance technologies based on real world crash data.Proceedings of the Future Active Safety Technology Towards ZeroTraffic Accidents (FASTzero), Gothenburg, Sweden. p. 9–11.
11. Cicchino JB. 2017. Effectiveness of forward collision warning and auton-omous emergency braking systems in reducing front-to-rear crashrates. Accid Anal Prev. 99(Pt A):142–152. doi:10.1016/j.aap.2016.11.009.
12. Cicchino JB. 2018. Effects of lane departure warning on police-reported crash rates. J Safety Res. 66:61–70. doi:10.1016/j.jsr.2018.05.006.

13. Teoh ER. 2022. Motorcycle antilock braking systems and fatal crash rates: updated results. Traffic Inj Prev. 23(4):203–207. doi:10.1080/15389588.2022.2047957.

14. Cummings M. 2024. Assessing readiness of self-driving vehicles. In:The 103rd Transportation Research Board (TRB) Annual Meeting.Washington, D.C.

15. Chen JJ, Shladover SE. 2024. Initial Indications of Safety of Driverless Automated Driving Systems. arXiv preprint arXiv:2403.14648.

16. Di Lillo L, Gode T, Zhou X, Atzei M, Chen R, Victor T. 2024. Comparative safety performance of autonomous-and human drivers: a real-world case study of the Waymo one service. Heliyon.10(14):e34379. doi:10.1016/j.heliyon.2024.e34379.

17. Di Lillo L, Gode T, Zhou X, Scanlon JM, Chen R, Victor T. 2025. Do autonomous vehicles outperform latest-generation human-driven vehicles? a comparison to waymo's auto liability insurance claims at25.3M miles. Mountain View (CA): Waymo LLC.

18. Kusano, Kristofer D., John M. Scanlon, Yin-Hsiu Chen, Timothy L. McMurry, Ruoshu Chen, Tilia Gode, and Trent Victor. 2024. "Comparison of Waymo Rider-Only Crash Data to Human Benchmarks at 7.1 Million Miles." Traffic Injury Prevention 25 (sup1): S66–77. doi:10.1080/15389588.2024.2380786.

19. Kusano, Kristofer D., John M. Scanlon, Yin-Hsiu Chen, Timothy L. McMurry, Tilia Gode, and Trent Victor. 2025. "Comparison of Waymo Rider-Only Crash Rates by Crash Type to Human Benchmarks at 56.7 Million Miles." Traffic Injury Prevention, May, 1–13. doi:10.1080/15389588.2025.2499887.

20. Banerjee S, Jha S, Cyriac J, Kalbarczyk ZT, Iyer RK. 2018. Hands off the wheel in autonomous vehicles?: a systems perspective on over a million miles of field data. In: 2018 48th Annual IEEE/IFIPInternational Conference on Dependable Systems and Networks(DSN), IEEE. p. 586–597. doi:10.1109/DSN.2018.00066.

21. Blanco M, Atwood J, Russell S, Trimble T, McClafferty J, Perez M. 2016.Automated vehicle crash rate comparison using naturalistic data.Final report. Blacksburg, VA: Virginia Tech Transportation Institute.

22. Hankey, Jonathan M., Miguel A. Perez, and Julie A. McClafferty. Description of the SHRP 2 naturalistic database and the crash, near-crash, and baseline data sets. Virginia Tech Transportation Institute, 2016.

23. Dixit V, Chand S, Nair D. 2016. Autonomous vehicles: disengagements, accidents and reaction times. PLoS One. 11(12):e0168,054. doi:10.1371/journal.pone.0168054.

24. Favarò F, Nader N, Eurich S, Tripp M, Varadaraju N. 2017. Examiningaccident reports involving autonomous vehicles in california. PLoSOne. 12(9):e0184,952. doi:10.1371/journal.pone.0184952.

25. Schoettle B, Sivak M. 2015. A preliminary analysis of real-world crash-es involving self-driving vehicles. Report no. UMTRI-2015-34. AnnArbor, MI: University of Michigan Transportation Research Institute.

26. Goodall, Noah J. "Comparison of automated vehicle struck-from-behind crash rates with national rates using naturalistic data." Accident Analysis & Prevention 154 (2021): 106056, https://doi.org/10.1016/j.aap.2021.106056.

27. Goodall NJ. 2021. Potential crash rate benchmarks for automated vehicles. Transp Res Rec. 2675(10):31–40. doi:10.1177/03611981211009878.

28. Teoh ER, Kidd DG. 2017. Rage against the machine? Google's self-driving cars versus human drivers. J Safety Res. 63:57–60. doi:10.1016/j.jsr.2017.08.008.

29. Cooper, D., et al., TNCs 2020: A Profile of Ride-Hailing in California, SFCTA, 2023, San Francisco County.

30. Flannagan C, Leslie A, Kiefer R, Bogard S, Chi-Johnston G, Freeman L, Huang R, Walsh D, Joseph A. 2023. Establishing a crash rate benchmark using large-scale naturalistic human ridehail data. Ann Arbor (MI):University of Michigan Transportation Research Institute (UMTRI).

31. Teoh, Eric R., and David G. Kidd. "Rage against the machine? Google's self-driving cars versus human drivers." Journal of safety research 63 (2017): 57-60.

32. Scanlon, John M., Kristofer D. Kusano, Laura A. Fraade-Blanar, Timothy L. McMurry, Yin-Hsiu Chen, and Trent Victor. 2024. "Benchmarks for Retrospective Automated Driving System Crash Rate Analysis Using Police-Reported Crash Data." Traffic Injury Prevention 25 (sup1): S51–65. doi:10.1080/15389588.2024.2380522.

33. Chen YH, Scanlon JM, Kusano KD, McMurry TL, Victor T. 2024. Dynamic benchmarks: spatial and temporal alignment for ADS performance eval-uation. arXiv preprint arXiv:2410.08903. doi:10.48550/arXiv.2410.08903.

34. Federal Highway Administration (FHWA). "Highway functional classification concepts, criteria and procedures 2023 Edition" (2023). https://rosap.ntl.bts.gov/view/dot/72430/dot_72430_DS1.pdf

35. American Association of State Highway and Transportation Officials, *A Policy on Geometric Design of Highways and Streets, 7th Edition*, American Association of State Highway and Transportation Officials: Washington, D.C. (2018), isbn: 978-1-56051-676-7.

36. Bareiss, Max, John Scanlon, Rini Sherony, and Hampton C. Gabler. "Crash and injury prevention estimates for intersection driver assistance systems in left turn across path/opposite direction crashes in the United States." Traffic injury prevention 20, no. sup1 (2019): S133-S138.

37. Scanlon, John M., Rini Sherony, and Hampton C. Gabler. "Injury mitigation estimates for an intersection driver assistance system in straight crossing path crashes in the United States." Traffic injury prevention 18, no. sup1 (2017): S9-S17.

38. Federal Highway Administration. 2023b. Highway statistics 2021. https://www.fhwa.dot.gov/policyinformation/statistics/2021/.

39. National Center for Statistics and Analysis. 2023b. Fatality analysis reporting system analytical user's manual, 1975-2021. Tech. Rep. DOTHS 813 417, National Highway Traffic Safety Administration.

40. Arizona Department of Transportation. 2023a. 2022 Motor Vehicle Crash Facts for the State of Arizona. Phoenix (AZ): the Arizona Department of Transportation.

41. Arizona Department of Transportation. 2023c. Records center. https://azdot.govqa.us/WEBAPP/_rs/supporthome.aspx.

42. Arizona Department of Transportation. 2023b. Extent and travel dash-board. https://experience.arcgis.com/experience/ac0948fc05224aa8a80313f59a634fde.

43. California Highway Patrol (CHP). 2024. California Crash ReportingSystem (CCRS) – Dataset – California Open Data Portal. CaliforniaHighway Patrol; 2024 May 31 [updated 2025 Feb 3]. https://data.ca.gov/dataset/ccrs.

44. California Department of Transportation (Caltrans). 2023. California Public Road Data: statistical Information Derived from the Highway Performance Monitoring System. California Department of Transportation.

45. Georgia Department of Transportation, "Crash Reporting," Georgia Department of Transportation, accessed July 2025, https://www.dot.ga.gov/GDOT/pages/CrashReporting.aspx.

46. Georgia Department of Transportation, "Road & Traffic Data," Georgia Department of Transportation, accessed July 2025, https://www.dot.ga.gov/GDOT/Pages/RoadTrafficData.aspx.

47. Texas Department of Transportation (TXDOT). 2022. Motor Vehicle Traffic Crash Data. Texas Department of Transportation [accessed 2023 Aug 15]. https://www.txdot.gov/apps-cg/crash_records/form.htm.

48. Texas Department of Transportation (TXDOT). 2023. 2023 Texas Transit Statistics. TXDOT Public Transportation Division

49. Federal Highway Administration. Highway Performance Monitoring System Field Manual. https://www.fhwa.dot.gov/policyinformation/hpms/fieldmanual/page01.cfm, 2018. Accessed Jul. 22,2024.

50. "Vehicle Identification Number (VIN) Requirements." Code of Federal Regulations, title 49, part 565 (2025). https://www.ecfr.gov/current/title-49/subtitle-B/chapter-V/part-565.

51. Kusano KD, Scanlon JM, Brännström M, Engström J, Victor T. 2023.Framework for a conflict typology including contributing factors foruse in ADS safety evaluation. Proceeding of 27th InternationalTechnical Conference on the Enhanced Safety of Vehicles. PaperNumber 23-0328.

52. Najm WG, Smith DL. 2007. Definition of a pre-crash scenario typology forvehicle safety research. Proceeding of 27th International TechnicalConference on the Enhanced Safety of Vehicles. Paper Number 07-0412.

53. Blincoe L, Miller T, Wang JS, Swedler D, Coughlin T, Lawrence B, GuoF, Klauer S, Dingus T. 2023. The economic and societal impact of motor vehicle crashes, 2019 (revised). Tech. Rep. DOT HS 813 403, National Highway Traffic Safety Administration.

54. Governors Highway Safety Association and National Highway Traffic Safety Administration. 2017. Mmucc guideline model minimum uniform crash criteria, 5th edition. Tech. Rep. DOT HS 812 433. Available at www.nhtsa.gov/mmucc-1.

55. Portney, Leslie Gross, and Mary P. Watkins. Foundations of clinical research: applications to practice. Vol. 892. Upper Saddle River, NJ: Pearson/Prentice Hall, 2009.

56. San Francisco Department of Public Health-Program on Health, Equityand Sustainability. 2017. Vision zero high injury network: 2017update – a methodology for San Francisco, California. Tech. rep.

57. American Association of State Highway and Transportation Officials, Highway Safety Manual (Washington, D.C.: AASHTO, 2010).

58. Bergel-Hayat, Ruth, Mohammed Debbarh, Constantinos Antoniou, and George Yannis. "Explaining the road accident risk: Weather effects." Accident Analysis & Prevention 60 (2013): 456-465.

59. Brumbelow, Matthew L., and Jessica S. Jermakian. 2021. "Injury Risks and Crashworthiness Benefits for Females and Males: Which Differences Are Physiological?" Traffic Injury Prevention 23 (1): 11–16. doi:10.1080/15389588.2021.2004312.

60. Farmer, C. M. (2019). The effects of higher speed limits on traffic fatalities in the United States, 1993–2017. Insurance Institute for Highway Safety.

61. Hu, Wen, and Jessica B. Cicchino. "Effects of lowering speed limits on crash severity in Seattle." Journal of safety research 88 (2024): 174-178.

62. Li G, Baker SP, Langlois JA, Kelen GD. Are female drivers safer? An application of the decomposition method. Epidemiology. 1998 Jul;9(4):379-84. PMID: 9647900.

63. National Center for Statistics and Analysis. (2025, January). Seat belt use in 2024 – overall results (Traffic Safety Facts Research Note. Report No. DOT HS 813 682). National Highway Traffic Safety Administration.

64. Xu, Chengcheng, Pan Liu, Wei Wang, and Zhibin Li. "Evaluation of the impacts of traffic states on crash risks on freeways." Accident Analysis & Prevention 47 (2012): 162-171.

65. Yadav, Ankit Kumar, and Nagendra R. Velaga. "Alcohol-impaired driving in rural and urban road environments: Effect on speeding behaviour and crash probabilities." Accident Analysis & Prevention 140 (2020): 105512.

66. Johnson, N. S., & Gabler, H. C. (2015). Injury Outcome in Crashes with Guardrail End Terminals. Traffic Injury Prevention, 16(sup2), S103–S108. https://doi.org/10.1080/15389588.2015.1065976

67. Kusano, Kristofer D., and Hampton C. Gabler. 2014. "Comprehensive Target Populations for Current Active Safety Systems Using National Crash Databases." Traffic Injury Prevention 15 (7): 753–61. doi:10.1080/15389588.2013.871003.

68. Scanlon, J. M., Kusano, K. D., & Gabler, H. C. (2016). Lane Departure Warning and Prevention Systems in the U.S. Vehicle Fleet: Influence of Roadway Characteristics on Potential Safety Benefits. Transportation Research Record, 2559(1), 17-23. https://doi.org/10.3141/2559-03 (Original work published 2016)

69. Schubert, Angela, Stefan Babisch, John M. Scanlon, Eamon T. Campolettano, Robby Roessler, Thomas Unger, and Timothy L. McMurry. "Passenger and heavy vehicle collisions with pedestrians: assessment of injury mechanisms and risk." Accident Analysis & Prevention 190 (2023): 107139.

70. Schubert, Angela, Eamon T. Campolettano, John M. Scanlon, Thomas Unger, and Timothy L. McMurry. "A Mechanistic Approach to Modeling Omnidirectional Motorcyclist Injury Risk" Traffic Injury Prevention (Accepted Manuscript) (2025).

71. Engström, J., Liu, S. Y., DinparastDjadid, A., & Simoiu, C. (2024). Modeling road user response timing in naturalistic traffic conflicts: A surprise-based framework. Accident Analysis & Prevention, 198, 107460.

72. Horstmann, G. (2006). Latency and duration of the action interruption in surprise. Cognition & Emotion, 20(2), 241-267.

73. Johnson, Leif, Johan Engström, Aravinda Srinivasan, Ibrahim Öztürk, and Gustav Markkula. "Looking for an out: Affordances, uncertainty and collision avoidance behavior of human drivers." arXiv preprint arXiv:2505.14842 (2025).

74. Scanlon, John M., Kristofer D. Kusano, Tom Daniel, Christopher Alderson, Alexander Ogle, and Trent Victor. "Waymo simulated driving behavior in reconstructed fatal crashes within an autonomous vehicle operating domain." Accident Analysis & Prevention 163 (2021): 106454.

# Appendix

## Fatality Rates from FHWA 2023 Highway Statistics Series

Figure A1 shows annual fatality data compiled by the Federal Highway Administration's (FHWA) as a part of the highway statistics series [38]. Using 2023 national fatal statistics (from the Fatality Analysis Reporting System, FARS [39]) and associated vehicle miles traveled (VMT) estimates, the data shows that the fatality rate (count of fatally injured persons by VMT) was 2.7 times higher on surface streets (1.60 fatalities per 100M VMT) than freeways (0.59 fatalities per 100M VMT). It is notable that risk is not uniform across all roadway functional classes within freeway and surface street groups. Most notably, "local" roads (0.94 fatalities per 100M VMT) were 58% of the average fatality rates on surface streets.

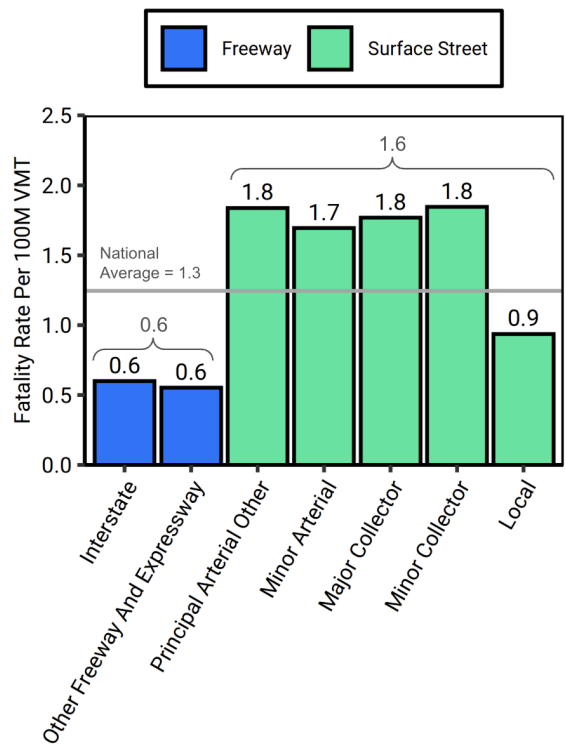

**Figure A1**. Fatality rates (persons fatally injured per 100M1B VMT) for all US roads in 2023 are shown for each FHWA functional class. All data is taken directly from the annual Highway Statistics Series. Fatality rates (count of all fatally injured persons in a crash) are not equivalent to crashed vehicle rates (count of only the vehicles involved in fatal crash rates), so *these rates should not be considered as an equivalent form to the crashed vehicle benchmark rates presented in this paper* [4, 32].

# In-transport Passenger Vehicle Type Classification

**Table A1.** Variables relied upon for selecting passenger vehicles in-tranport from the crash data.

|  |  | Variables used | |
| --- | --- | --- | --- |
| **State** | **Data Source** | **In-Transport** | **Passenger Vehicles** |
| **California** | **CHP SWITRS** | Move_pre_acc, Party_type | party_type, Stwd_vehicle_type, Chp_veh_type_towing |
| **Arizona** | **ADOT Crash Data** | UnitAction | BodyStyle, UnitType |
| **Georgia** | **GDOT GEARS** | mnvrveh | vehtype |
| **Texas** | **TXDOT CRIS** | Veh_Parked_Fl | VIN*, Veh_Body_Styl_ID, Cmv_Veh_Type_ID, Unit_Desc_ID, Cmv_GVWR, Cmv_Fiveton_Fl |

# Crash Type Classification

**Table A2**. Variables used to subset identify actor type and crash type

|  |  | Variables used | |
| --- | --- | --- | --- |
| **State** | **Data Source** | **Actor Type** | **Crash Type (in addition to variables in actor type)** |
| **Arizona** | **ADOT Crash Data** | UnitAction, BodyStyle, UnitType | party_count, EventSequence1, CollisionManner, JunctionRelation, UnitAction |
| **California** | **CHP SWITRS** | Party_type, Stwd_vehicle_type, Chp_veh_type_towing | party_number, pcf_violation, move_pre_acc, dir_of_travel, type_of_collision |
| **Georgia** | **GDOT GEARS** |  |  |
| **Texas** | **TXDOT CRIS** | Veh_Parked_Fl, VIN*, Veh_Body_Styl_ID, Cmv_Veh_Type_ID, | FHE_Collsn_ID, Contrib_Factr_1_ID, Contrib_Factr_2_ID, |

| | | Unit_Desc_ID, Cmv_GVWR, Cmv_Fiveton_Fl | Contrib_Factr_3_ID |
|---|---|---|---|